\documentclass[10pt,twocolumn,letterpaper]{article}

\usepackage[pagenumbers]{cvpr} 

\definecolor{cvprblue}{rgb}{0.21,0.49,0.74}
\usepackage[pagebackref,breaklinks,colorlinks,allcolors=cvprblue]{hyperref}

\def\paperID{9121} 
\def\confName{CVPR}
\def\confYear{2025}

\usepackage{algorithm}
\usepackage{algorithmicx}
\usepackage{algpseudocode}
\usepackage{setspace}
\usepackage{array}
\usepackage{multirow}

\usepackage{cuted}
\usepackage{capt-of}

\begin{document}
\title{Generative Densification: Learning to Densify Gaussians for High-Fidelity Generalizable 3D Reconstruction}

\newcommand\CoAuthorMark{\footnotemark[\arabic{footnote}]}

\author{
Seungtae Nam$^{1}$\thanks{Equal contribution}\hspace{2.2mm}
Xiangyu Sun$^{2}$\protect\CoAuthorMark\hspace{2.2mm}
Gyeongjin Kang$^{2}$\hspace{2.2mm}
Younggeun Lee$^{2}$\hspace{2.2mm}
Seungjun Oh$^{2}$\hspace{2.2mm}
Eunbyung Park$^{1}$\thanks{Corresponding author}
\vspace{2mm} \\
$^1$Yonsei University\hspace{2.2mm}$^2$Sungkyunkwan University
\vspace{2mm} \\
{\small \url{https://stnamjef.github.io/GenerativeDensification/}}
}

\maketitle

\begin{strip}
    \centering
    \vspace*{-16mm}
    \includegraphics[width=1.0\textwidth]{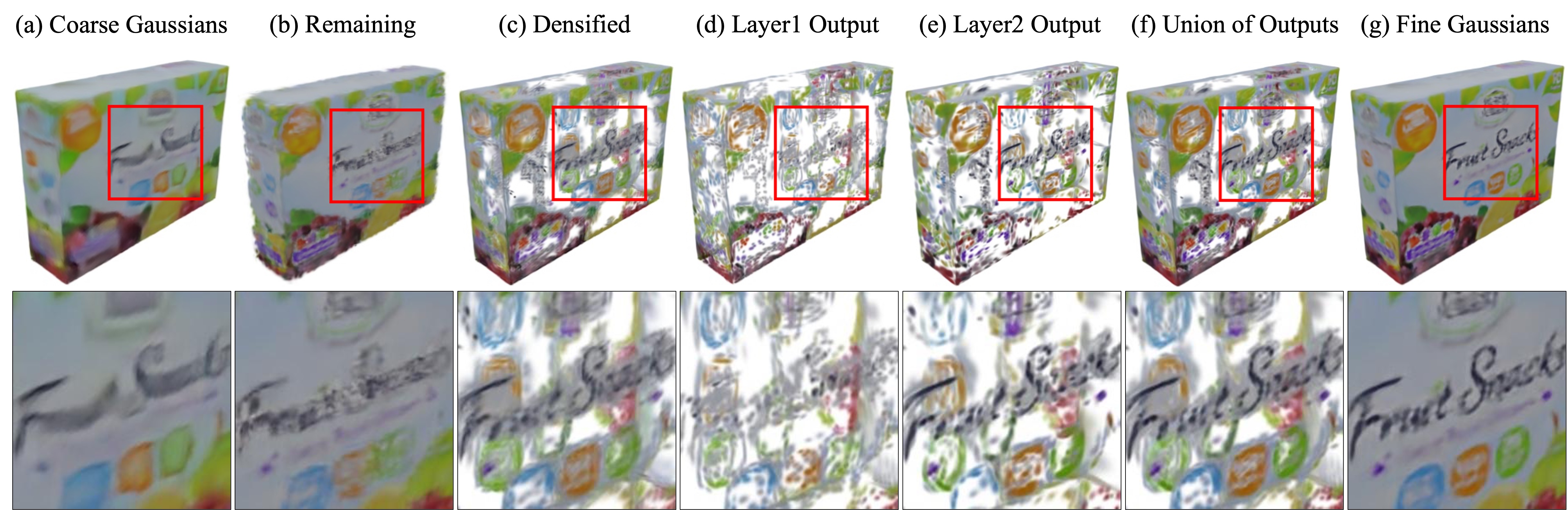}
    \vspace*{-6mm}
    \captionof{figure}{
    Our method selectively densifies (a) coarse Gaussians from generalized feed-forward models. 
    (c) The top $K$ Gaussians with large view-space positional gradients are selected, and (d-e) their fine Gaussians are generated in each densification layer. 
    (g) The final Gaussians are obtained by combining (b) the remaining (non-selected) Gaussians with (f) the union of each layer's output Gaussians.
    }
    \label{fig:main_result}
\end{strip}

\begin{abstract}
Generalized feed-forward Gaussian models have achieved significant progress in sparse-view 3D reconstruction by leveraging prior knowledge from large multi-view datasets.
However, these models often struggle to represent high-frequency details due to the limited number of Gaussians.
While the densification strategy used in per-scene 3D Gaussian splatting (3D-GS) optimization can be adapted to the feed-forward models, it may not be ideally suited for generalized scenarios.
In this paper, we propose Generative Densification, an efficient and generalizable method to densify Gaussians generated by feed-forward models.
Unlike the 3D-GS densification strategy, which iteratively splits and clones raw Gaussian parameters, our method up-samples feature representations from the feed-forward models and generates their corresponding fine Gaussians in a single forward pass, leveraging the embedded prior knowledge for enhanced generalization.
Experimental results on both object-level and scene-level reconstruction tasks demonstrate that our method outperforms state-of-the-art approaches with comparable or smaller model sizes, achieving notable improvements in representing fine details.
\end{abstract}

\begin{figure*}
    \centering
    \includegraphics[width=1.0\textwidth]{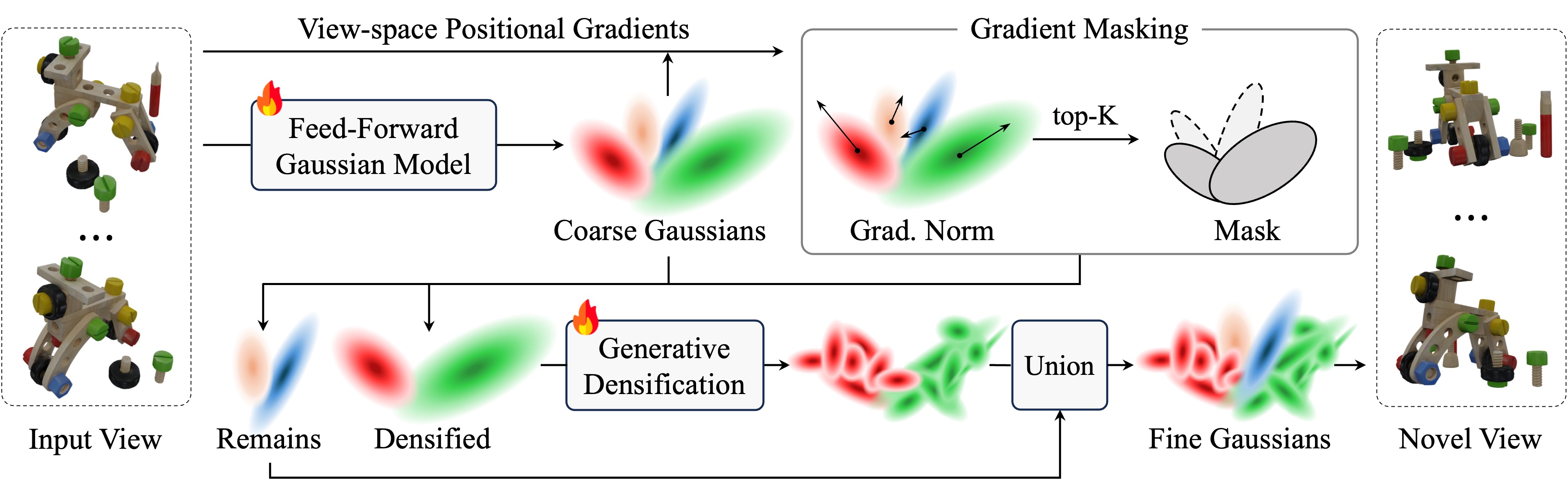}
    \vspace*{-6mm}
    \caption{
    Generative Densification overview.
    We selectively densifies the top $K$ Gaussians with large view-space positional gradients.
    }
    \vspace{-4mm}
    \label{fig:overview_arch}
\end{figure*}

\vspace{-7mm}
\section{Introduction}
\label{sec:introduction}

3D Gaussian splatting (3D-GS)~\cite{kerbl20233d} has been a massive success in high-quality 3D scene reconstruction and real-time novel view synthesis, representing a 3D scene with a set of learnable Gaussian primitives and exploiting a fast differentiable rasterization pipeline. 
Since its introduction, numerous studies have explored 3D-GS as a versatile 3D representation with various applications extending beyond the per-scene optimization. 
One notable example is generalized feed-forward Gaussian models~\cite{szymanowicz2024splatter, chen2025lara, tang2025lgm, xu2024grm, zhang2025gs-lrm, charatan2024pixelsplat, chen2025mvsplat, zheng2024gps, zhang2024geolrm}, which generate 3D Gaussian primitives to reconstruct 3D scenes or objects in a single forward pass. 
By leveraging 3D prior knowledge learned from large multi-view datasets (e.g., Objaverse~\cite{deitke2023objaverse, deitke2024objaverse}, RE10K~\cite{zhou2018stereo}), these models can reconstruct the 3D scene or objects with the generated 3D Gaussians from only a single or a few images. 

While effective and promising, the feed-forward Gaussian models often face challenges in capturing high-frequency details primarily due to the limited number of Gaussians and the lack of a densification strategy tailored for these approaches. 
One possible solution is to generate more Gaussians, but this does not fully address the problem. 
For example, in the pixel-aligned Gaussian model~\cite{charatan2024pixelsplat}, a widely adopted architecture in the feed-forward models (see \cref{sec:related_work} for details), we can predict multiple Gaussians per pixel.
However, this strategy uniformly increases the number of Gaussians across the entire 3D space, resulting in an excess of unnecessary small Gaussians that may degrade rendering quality and speed. 
Ideally, the scene details should be represented by numerous small Gaussians, while smoother regions can be covered by a few large Gaussians.

In per-scene 3D-GS optimization, an adaptive densification strategy selectively increases the number of Gaussians to better reconstruct complex 3D geometry and fine details.
Specifically, every few optimization steps, it densifies only the Gaussians with large view-space positional gradients, replicating the existing Gaussians or splitting them into smaller ones when more Gaussians are required. 
Through repeated optimization and densification processes, the Gaussians gradually become non-uniformly distributed, with numerous small Gaussians concentrated in detailed areas and a few large Gaussians scattered across smooth regions. 
This simple heuristic algorithm has demonstrated its effectiveness across various real-world scenes and objects.

A straightforward approach to extend the densification strategy to the feed-forward Gaussian models is to fine-tune the output Gaussians from the feed-forward models using the original 3D-GS optimization and densification procedure. 
However, directly applying the densification strategy used in per-scene optimization scenarios presents several challenges. 
For example, it demands hundreds and thousands of optimization steps to converge and reconstruct the fine details, which significantly reduces generation speed. 
Additionally, the per-scene 3D-GS optimization usually assumes many input images from various viewpoints, whereas the feed-forward models typically take just a few views or even a single view image. 
Thus, the densification and extensive optimization to a few view images can easily lead to overfitting, compromising the feed-forward models’ ability to generalize and synthesize novel views effectively.

In this paper, we propose \textit{Generative Densification (GD)}, an efficient generative densification strategy for generalized feed-forward Gaussian models, which is designed in accordance with the following principles: 1) adaptability, allowing the model to distinguish Gaussians that requires densification and those do not; and 2) generalizability, enabling the model to learn and leverage prior knowledge from large multi-view datasets. 
Inspired by the 3D-GS densification strategy, our approach leverages view-space positional gradients to identify where additional Gaussians are needed. 
We selectively densify the top $K$ Gaussians with large gradients, and the remaining (non-selected) Gaussians are used together with the densified ones to render high-fidelity images (\cref{fig:overview_arch}).
Unlike the 3D-GS densification strategy, the proposed \textit{GD} densifies the selected Gaussians based on the learned prior knowledge for the purpose of generalizable 3D reconstruction. 
More specifically, \textit{GD} densifies the feature representations from the feed-forward models rather than the actual Gaussians, and the prior knowledge is embedded in these features for better generalization.

We further propose to utilize an efficient point-level transformer~\cite{wu2024point} to implement \textit{GD}. 
Due to the quadratic increase in memory and the unstructured nature of point-level 3D data, it is infeasible to apply naive self-attention on hundreds of thousands of input Gaussians. 
While local attention can be applied within groups of neighboring Gaussians, finding the neighbors by calculating and comparing distances between the Gaussians is computationally expensive. 
Instead, we sort the Gaussians in traversing order of space-filling curves~\cite{peano1990courbe} to rearrange them into non-overlapping groups and apply group-wise attention, enabling efficient operations directly in 3D space. 
Additionally, while we selectively densify the Gaussians with large view-space positional gradients, multiple rounds of densification are unnecessary for all of them. 
To optimize this process, we predict a confidence mask for each densification layer to filter out the Gaussians that do not require further densification.

We integrated \textit{GD} into two recent feed-forward Gaussian models: LaRa~\cite{chen2025lara} for object-level reconstruction and MVSplat~\cite{chen2025mvsplat} for scene-level reconstruction.
On the large-scale Gobjaverse~\cite{xu-gobjaverse} and RE10K~\cite{zhou2018stereo} datasets, two models incorporating our method achieved the best performance, significantly improving reconstruction quality compared to their respective baselines.
Qualitative analysis further highlights that the fine Gaussians generated by \textit{GD} effectively capture thin structures and intricate details, which are often challenging for the coarse Gaussians to represent accurately (\cref{fig:main_result}).
Additionally, cross-dataset evaluations confirm that \textit{GD} consistently improves image quality, demonstrating its robust generalizability across different datasets.

\section{Related Work}
\label{sec:related_work}

\paragraph{Feed-forward Gaussian Models.}
Feed-forward Gaussian models learn a mapping from a single or a set of few input images to a set of Gaussian representations that can be rendered into any viewpoint. 
One common approach combines a pre-trained image encoder (e.g., DINO~\cite{caron2021emerging,oquab2023dinov2}) with a learnable embedding decoder, where the decoder generates explicit 3D features via a series of cross-attention between 3D embeddings and image features. 
These 3D features are commonly represented as voxels (LaRa~\cite{chen2025lara} and Geo-LRM~\cite{zhang2024geolrm}), point-cloud (Point-to-Gaussian\cite{lu2024large-point-to-gaussian}), or a hybrid of point-cloud and triplane (Triplane Gaussian Splatting~\cite{zou2024triplane-gs}), and the Gaussian parameters can be obtained by decoding the features using an MLP. 
Covering a wide range of view with an explicit 3D representation, this approach is effective for 360-degree object-level reconstruction. 
However, the cross-attention on the 3D features is memory-intensive and computationally expensive.

Another approach involves pixel-aligned Gaussian representations~\cite{szymanowicz2024splatter, charatan2024pixelsplat}, where each Gaussian is assumed to be on a ray of each pixel and distanced from the ray origin by a depth. 
The depth is either directly regressed from pixel-wise image features~\cite{szymanowicz2024splatter} or determined by the probability that a Gaussian exists at a depth along a ray~\cite{charatan2024pixelsplat}. 
Since its introduction, substantial efforts have been made to estimate the depth more accurately and extend it to large models for better generalizability. 
MVSplat~\cite{chen2025mvsplat} and MVSGaussian~\cite{liu2025mvsgaussian} propose to utilize cost volumes to leverage cross-view feature similarities for depth estimation, and, more recently, Flash3D~\cite{szymanowicz2024flash3d} and DepthSplat~\cite{xu2024depthsplat} further use a pre-trained depth estimator, enabling more robust estimation.
LGM~\cite{tang2025lgm} and other works~\cite{xu2024grm, zhang2025gs-lrm} extend the pixel-aligned Gaussians to the large models by attaching a Gaussian head layer on the top of 2D U-Net or ViT. 
Trained on large multi-view datasets, these models can generate the Gaussians even when out-of-domain input images (e.g., generated images from text-to-3D models~\cite{shi2023mvdream}) are given.

While both of these approaches can effectively generate Gaussian representations from a single or a few input images, they often struggle to reconstruct fine details. 
This is primarily due to the limited number of Gaussians to represent the details and the lack of a densification strategy to refine the Gaussians generated by the feed-forward models.

\vspace{-3mm}
\paragraph{Adaptive Density Control of Gaussians.}
The adaptive densification strategy~\cite{kerbl20233d} plays a crucial role in filling empty areas and capturing fine details.
It involves calculating the norm of view-space positional gradients averaged across different views, with only Gaussians that have large norms being selected for densification.
While effective, the algorithm depends on having good initial Gaussian positions, and large Gaussians are often not split into smaller ones, leading to blurry images in the final renderings.
MCMC-GS~\cite{kheradmand20243d} proposes to optimize the Gaussians with Stochastic Gradient Langevin Dynamics update, improving the robustness to initialization.
To address the issue of large Gaussians, AbsGS~\cite{ye2024absgs} introduces homodirectional view-space gradients as a criterion of densification, while other methods~\cite{zhang2024pixel, bulo2024revising} consider the number of pixels covered by each Gaussian when calculating the gradients.

Although the methods outlined above successfully improve the rendering quality, they all require more than thousands of optimization and densification steps, and the discussions are limited to per-scene optimization scenarios.
Our goal is to develop an efficient densification method tailored for generalized feed-forward Gaussian models.
Instead of directly applying the existing methods in generalized settings, we learn from large multi-view datasets to generate fine Gaussians in a single forward pass.
We show that our method can be generalized to a variety of objects and scenes by leveraging the learned prior knowledge.

\section{Method}
\label{sec:method}

\subsection{Generative Densification}
\label{subsec:generative densification}

A feed-forward Gaussian model is a function $\Phi$ that maps a set of images $\mathcal{I}=\{I_v\}_{v=1}^{V}$ and camera poses $\mathcal{C}=\{C_v\}_{v=1}^{V}$ to a set of Gaussians and features,
\begin{equation}
\label{eq:eq1}
    \mathcal{G}^{(0)}, \mathcal{F}^{(0)} = \Phi(\mathcal{I}, \mathcal{C}),
\end{equation}
where $V$ is the number of input views, and $\mathcal{F}^{(0)}$ is the features that are used to generate the Gaussians (see \cref{subsec:applying_generative_densification}). Each Gaussian consists of positions, an opacity, spherical harmonics (SH) coefficients, quaternions, and scales. 

Our goal is to learn a densification model $\Psi$ that can adaptively densify Gaussians generated by feed-forward models for high-fidelity 3D reconstruction:
\begin{equation}
\label{eq:eq2}
    \hat{\mathcal{G}} = \Psi(\mathcal{G}^{(0)}, \mathcal{F}^{(0)}, \mathcal{I}, \mathcal{C}).
\end{equation}
First, using the view-space positional gradients, we split the Gaussians into two groups: the Gaussians requiring densification ($\mathcal{G}_\text{den}^{(0)}$) and the remaining Gaussians ($\mathcal{G}_{\text{rem}}^{(0)}$).
Specifically, we compute scores as the norms of gradients with respect to the view-space (or projected) coordinates of Gaussian positions, averaged across the $V$ input views:
\begin{equation}
\label{eq:eq3}
m_i^{(0)} = \frac{1}{V} \sum_{v=1}^V \lVert \nabla_{p(x_i^{(0)},v)} \mathcal{L}_\text{MSE}(I_v, \hat{I}_v) \rVert_2,
\end{equation}
where $p(x_i^{(0)},v) \in \mathbb{R}^2$ denotes the projected coordinate of $i$-th Gaussian position $x_i^{(0)} \in \mathbb{R}^3$ onto input view $v$, and $\mathcal{L}_\text{MSE}$ is the mean squared error between ground truth input images ($I_v$) and the rendered images ($\hat{I}_v$).
Based on the computed scores, the top $K^{(0)}$ scoring Gaussians ($\mathcal{G}_\text{den}^{(0)}$) are selected for densification.

\begin{figure}
    \centering
    \includegraphics[width=1.0\linewidth]{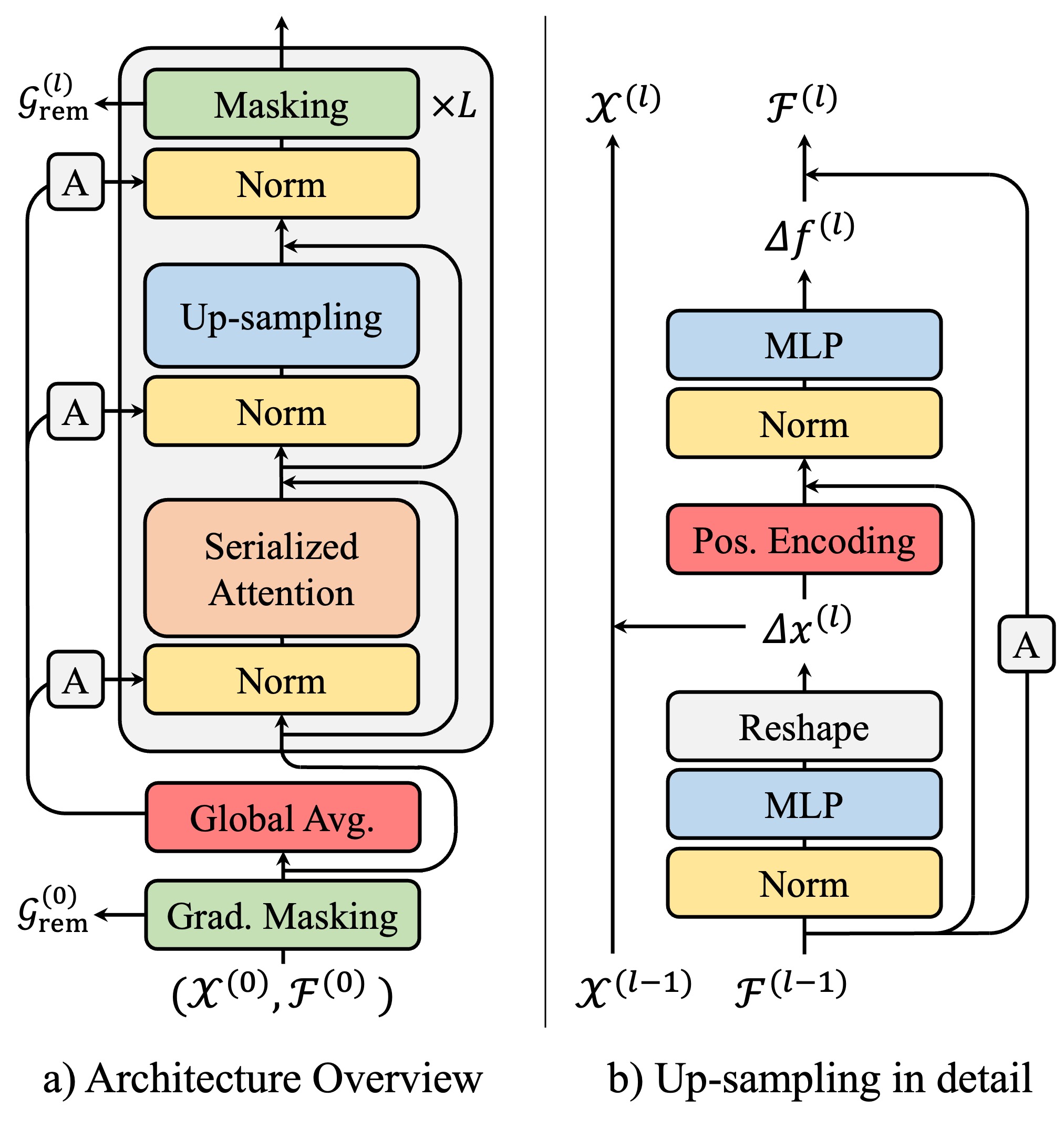}
    \vspace*{-8mm}
    \caption{
    Key components in Generative Densification Module.
    }
    \vspace{-5mm}
    \label{fig:overview_gdm}
\end{figure}

Then, the positions and features ($\mathcal{X}_\text{den}^{(0)} \in \mathbb{R}^{K^{(0)} \times 3}$, $\mathcal{F}_\text{den}^{(0)} \in \mathbb{R}^{K^{(0)} \times C}$) of the selected Gaussians are passed to the densification module (\cref{fig:overview_gdm}), which consists of up-sampling ($\texttt{UP}$), splitting via learnable masking (\texttt{SPLIT}), and Gaussian head ($\texttt{HEAD}$) components:
\begin{align}
    (\mathcal{X}^{(l)}, \mathcal{F}^{(l)}) &= \texttt{UP}(\mathcal{X}_\text{den}^{(l-1)}, \mathcal{F}_\text{den}^{(l-1)}), \label{eq:eq4} \\
    (\mathcal{X}_\text{den}^{(l)}, \mathcal{F}_\text{den}^{(l)}, \mathcal{X}_\text{rem}^{(l)}, \mathcal{F}_\text{rem}^{(l)}) &= \texttt{SPLIT}(\mathcal{X}^{(l)}, \mathcal{F}^{(l)}), \label{eq:eq5} \\
    \mathcal{G}^{(l)} &= \texttt{HEAD}(\mathcal{X}_\text{rem}^{(l)}, \mathcal{F}_\text{rem}^{(l)}) \label{eq:eq6},
\end{align}
for $l \in \{1, {\cdots}, L{-}1\}$.
$\texttt{UP}(\cdot, \cdot)\colon \mathbb{R}^{K^{(l-1)} \times 3} \times \mathbb{R}^{K^{(l-1)} \times C} \rightarrow \mathbb{R}^{K^{(l)} \times 3} \times \mathbb{R}^{K^{(l)} \times C}$, up-samples the Gaussian positions and features.
$\texttt{SPLIT}(\cdot, \cdot)\colon \mathbb{R}^{K^{(l)} \times 3} \times \mathbb{R}^{K^{(l)} \times C} \rightarrow \mathbb{R}^{K^{(l)}_\text{den} \times 3} \times \mathbb{R}^{K^{(l)}_\text{den} \times C} \times \mathbb{R}^{K^{(l)}_\text{rem} \times 3} \times \mathbb{R}^{K^{(l)}_\text{rem} \times C} \ (K^{(l)} = K^{(l)}_\text{den} + K^{(l)}_\text{rem})$, divides the up-sampled positions and features into those requiring further densification in the next layer ($\mathcal{X}_\text{den}^{(l)}, \mathcal{F}_\text{den}^{(l)}$) and the remaining ones ($\mathcal{X}_\text{rem}^{(l)}, \mathcal{F}_\text{rem}^{(l)}$).
More details on the number of Gaussians after the up-sampling and splitting operations are provided in \cref{sec:suppl_model_details}.
Using the remaining Gaussian positions and features, fine Gaussians for each layer ($\mathcal{G}^{(l)}$) are generated via the $\texttt{HEAD}(\cdot, \cdot)$ module.
Finally, the $L$-th layer's fine Gaussians are generated as $\mathcal{G}^{(L)} = \texttt{HEAD}(\texttt{UP}(\mathcal{X}_\text{den}^{(L-1)}, \mathcal{F}_\text{den}^{(L-1)}))$, and the final set of Gaussians is obtained as follows:
\begin{equation}
\label{eq:eq7}
    \hat{\mathcal{G}} = \mathcal{G}_\text{rem}^{(0)} \cup \{ \bigcup_{l=1}^{L} \mathcal{G}^{(l)} \}.
\end{equation}

\begin{figure*}[t]
    \centering
    \includegraphics[width=1.0\textwidth]{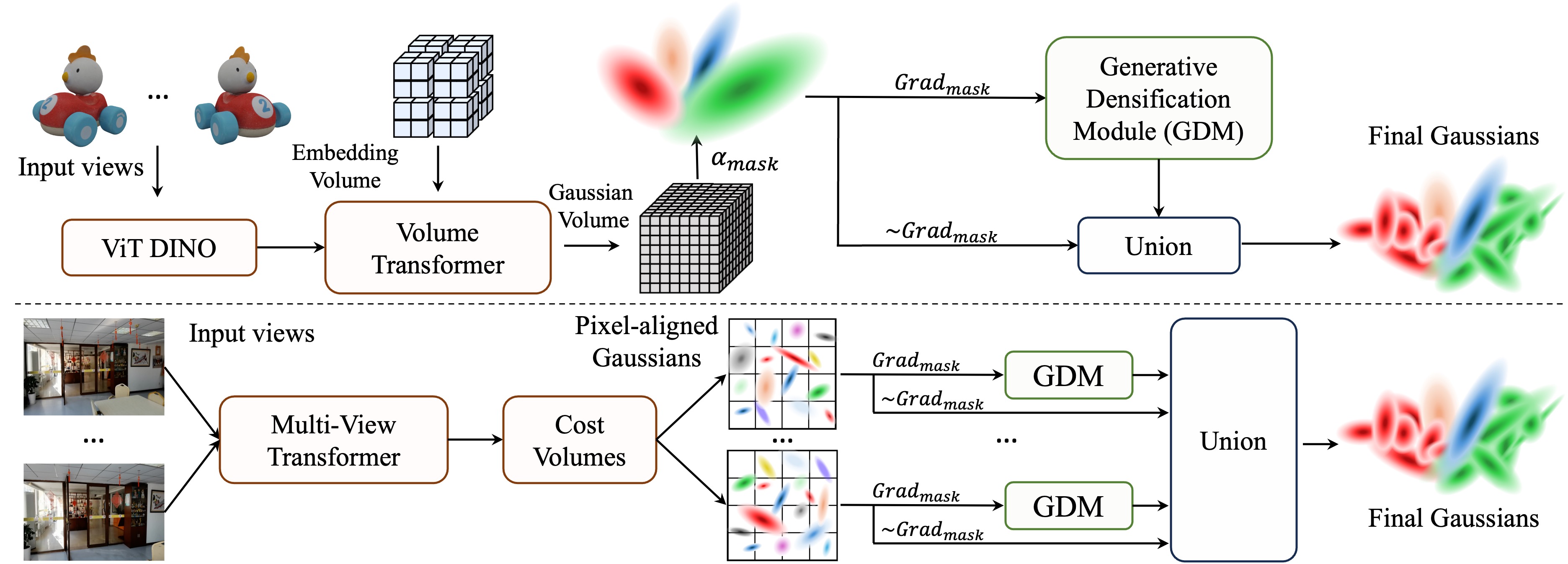}
    \vspace*{-6mm}
    \caption{
    Overview of the Generative Densification pipelines for object-level (top) and scene-level (bottom) reconstruction tasks.
    }
    \vspace{-1.25em}
    \label{fig:pipeline_object_scene}
\end{figure*}

\subsection{Architecture Details}
\label{subsec:architecture_details}

\paragraph{Serialized Attention.}
Due to the quadratic increase in memory requirements and the unstructured nature of the Gaussian representations, it is infeasible to apply self-attention directly to all input Gaussians.
One alternative is to group the Gaussians by searching for their neighbors and apply group-wise attention.
While this approach allows the model to consider the relative positions of the neighbors, searching for neighbors across hundreds of thousands of Gaussians is memory and computationally expensive.

Serialized attention~\cite{wu2024point} enables efficient operations on unstructured point clouds by sorting them in traversing order of given space-filling curves~\cite{peano1990courbe}.
Following its design principles, we first encode each Gaussian position into a serialized code, an integer value that reflects its order in the space-filling curves~\cite{wu2024point}.
Then, the Gaussian features are structured by sorting the serialized codes and divided into non-overlapping groups, where self-attention is applied within the same group.
By adapting serialized attention, our method efficiently embeds prior knowledge and local scene context into the features, allowing the output features to be used as sources for generating fine Gaussians.

\vspace{-4mm}
\paragraph{Up-sampling.}
In the $\texttt{UP}$ module, we predict residuals for each Gaussian position and feature (\cref{fig:overview_gdm}).
We generate $R^{(l)}$ offsets for each Gaussian position by passing the input features through an MLP parameterized by $\theta^{(l)}_{x}$.
These predicted offsets are then positionally encoded, concatenated with the input features, and transformed to residual features using another MLP parameterized by $\theta^{(l)}_{f}$. More formally,
\begin{align}
    {\Delta}x^{(l)}_i &= \texttt{MLP}({f}^{(l-1)}_i; \theta^{(l)}_{x}), \label{eq:eq8}\\
    {\Delta}f^{(l)}_{i,j} &= \texttt{MLP}(\gamma({\Delta}x^{(l)}_{i,j}) \oplus {f}^{(l-1)}_i; \theta^{(l)}_{f}),\label{eq:eq9}
\end{align}
where ${f}^{(l-1)}_i \in \mathbb{R}^{C}$ is the feature of $i$-th Gaussian after the serialized attention module.
${\Delta}x^{(l)}_i\in\mathbb{R}^{R^{(l)}{\times}3}$ is the predicted $R^{(l)}$ offsets per Gaussian, and ${\Delta}x^{(l)}_{i, j} \in\mathbb{R}^3$ denotes each predicted offset.
Similarly, ${\Delta}f^{(l)}_i\in\mathbb{R}^{R^{(l)}{\times}C}$ is the predicted $R^{(l)}$ residuals per Gaussian feature, and ${\Delta}f^{(l)}_{i,j}\in\mathbb{R}^{C}$ denotes each predicted residual. $\gamma(\cdot)$ is a positional encoding function, and $\oplus$ represents a concatenation operator.
The output positions and features are obtained by adding the residuals to their respective inputs.

\vspace{-3mm}
\paragraph{Splitting via Learnable Masking.}
In the $\texttt{SPLIT}$ module, we introduce learnable masking to filter out Gaussians not requiring further densification.
Although we use gradient masking to identify the Gaussians that need refinement in the first layer, calculating gradients at every layer incurs considerable computational overhead.
Instead, we predict a confidence score for each Gaussian using an MLP parameterized by $\theta^{(l)}_m$ and a sigmoid function $\sigma(\cdot)$.
More formally,
\begin{equation}
\label{eq:eq10}
m^{(l)}_i = \sigma(\texttt{MLP}(f^{(l)}_i ;\theta^{(l)}_m)),
\end{equation}
where $f^{(l)}_i \in \mathbb{R}^{C}$ is the feature for $i$-th Gaussian. 
Based on these predicted scores, the top $K^{(l)}_\text{den}$ scoring Gaussians are selected for further densification, while the remaining $K^{(l)}_\text{rem}$ Gaussians are decoded into fine Gaussian parameters.

Since selecting the $K^{(l)}_\text{den}$ Gaussians is non-differentiable, we attach a computational graph of $m_i^{(l)} \in [0,1]$ to input features $f_i^{(l)}$ while keeping the feature values intact:
\begin{equation}
\label{eq:eq11}
    f_i^{(l)} = \texttt{sg}(f_i^{(l)} - f_i^{(l)} m^{(l)}_i) + f_i^{(l)} m^{(l)}_i,
\end{equation}
where $\texttt{sg}(\cdot)$ is a stop-gradient operator.
Analogous to the straight-through estimator~\cite{yin2019understanding}, the gradients with respect to the output features $\partial{\mathcal{L}}/\partial{f_i^{(l)}}$ propagate back to the second term of the ~\cref{eq:eq13}, hence, which enables us to learn the masks and do end-to-end training.

\vspace{-3mm}
\paragraph{Gaussian Head.}
In the Gaussian $\texttt{HEAD}$ module, the positions and features of the remaining Gaussians ($\mathcal{X}_\text{rem}^{(l)}, \mathcal{F}_\text{rem}^{(l)}$), filtered out by the learnable masking module, are decoded into fine Gaussians ($\mathcal{G}^{(l)}$).
The positions of the fine Gaussians are set to be the same as the input positions, while the remaining attributes (opacities, SH coefficients, quaternions, and scales) are generated from the input features using an MLP.
Sigmoid and exponential functions are applied to the MLP outputs for opacities and scales, respectively, and the quaternions are normalized following~\cite{kerbl20233d}.

\vspace{-3.5mm}
\paragraph{Global Adaptive Normalization.}
The serialized attention module is learned to aggregate the scene context but operates within each group of Gaussians for memory and computational efficiency, which may lead to limited understanding of the global context.
To complement the local features, a global feature is widely used as a global descriptor in point-level architectures.
Inspired by previous works~\cite{qi2017pointnet, xiang2021snowflakenet} and recent normalization techniques~\cite{xu2019understanding, peebles2023scalable}, we introduce global adaptive normalization, which averages the features of the Gaussians selected for densifcation ($\mathcal{F}^{(0)}_\text{den}$) and scales the normalized features using the averaged features.

\subsection{Applying Generative Densification}
\label{subsec:applying_generative_densification}
We present two models that incorporates generative densification, based on LaRa~\cite{chen2025lara} for object-level reconstruction and MVSplat~\cite{chen2025mvsplat} for scene-level reconstruction.
The overall pipelines of generative densification for both scenarios are illustrated in \cref{fig:pipeline_object_scene}.
For object-level reconstruction, fine Gaussians are generated using the Gaussians and volume features produced by the LaRa backbone (top row of \cref{fig:pipeline_object_scene}).
For scene-level reconstruction, fine Gaussians are generated per view by utilizing the pixel-aligned Gaussians and image features extracted from the MVSplat backbone (bottom row of \cref{fig:pipeline_object_scene}).
Additionally, residual learning is incorporated into the scene-level model to better reconstruct complex indoor and outdoor real-world scenes.
Further details on the residual learning and the pipelines are provided in \cref{sec:suppl_generating_residuals} and \cref{sec:suppl_model_details}, respectively.

\begin{table*}
    \centering
    \caption{
    Quantitative comparisons of our object-level models against their baselines. 
    `Our-fast' is trained on the Gobjaverse~\cite{xu-gobjaverse} training set for 30 epochs, and `Ours' is further fine-tuned for 20 epochs.
    `Ours (w/ residual)' is trained on the same set for 50 epochs with residual learning (\cref{sec:suppl_generating_residuals}).
    LaRa is re-evaluated using the publicly available checkpoint and our view-sampling method (\cref{sec:suppl_training_and_evaluation_details}).
    }
    \vspace*{-3mm}
    \resizebox{\textwidth}{!}{
    \footnotesize
        \begin{tabular}{lcccccccccc}
        \toprule
        \multirow{2}{*}{Method} & \multirow{2}{*}{\#Param(M)} & \multicolumn{3}{c}{Gobjaverse~\cite{xu-gobjaverse}} & \multicolumn{3}{c}{GSO~\cite{downs2022google}} & \multicolumn{3}{c}{Co3D~\cite{reizenstein2021common}} \\
        \cmidrule(lr){3-5}\cmidrule(lr){6-8}\cmidrule(lr){9-11}
        & & PSNR $\uparrow$ & SSIM $\uparrow$ & LPIPS $\downarrow$ & PSNR $\uparrow$ & SSIM $\uparrow$ & LPIPS $\downarrow$ & PSNR $\uparrow$ & SSIM $\uparrow$ & LPIPS $\downarrow$ \\
        \midrule
        MVSNeRF~\cite{chen2021mvsnerf} & 0.52 & 14.48 & 0.896 & 0.185 & 15.21 & 0.912 & 0.154 & 12.94 & 0.841 & 0.241 \\
        MuRF~\cite{xu2024murf} & 15.7 & 14.05 & 0.877 & 0.301 & 12.89 & 0.885 & 0.279 & 11.60 & 0.815 & 0.393 \\
        \midrule
        LGM~\cite{tang2025lgm} & 415 & 19.67 & 0.867 & 0.157 & 23.67 & 0.917 & 0.063 & 13.81 & 0.739 & 0.414 \\
        GS-LRM~\cite{zhang2025gs-lrm} & 300 & - & - & - & 30.52 & 0.952 & \textbf{0.050} & - & - & - \\
        LaRa~\cite{chen2025lara} & 125 & 27.49 & 0.938 & 0.093 & 29.70 & 0.959 & 0.060 & 21.18 & 0.862 & 0.216 \\
        Ours-fast & 134 & 28.23 & 0.943 & 0.084 & 30.62 & 0.964 & 0.062 & 21.67 & 0.864 & 0.209 \\
        Ours & 134 & 28.58 & 0.945 & 0.080 & 31.06 & 0.966 & 0.058 & 21.72 & 0.865 & 0.209 \\
        Ours (w/ residual) & 134 & \textbf{28.75} & \textbf{0.946} & \textbf{0.078} & \textbf{31.23} & \textbf{0.967} & 0.058 & \textbf{22.08} & \textbf{0.867} & \textbf{0.206} \\
        \bottomrule
        \end{tabular}
    }
    \vspace{-3mm}
    \label{tab:quantitative result on object}
\end{table*}

\begin{figure*}[t]
    \centering
    \includegraphics[width=1.0\textwidth]{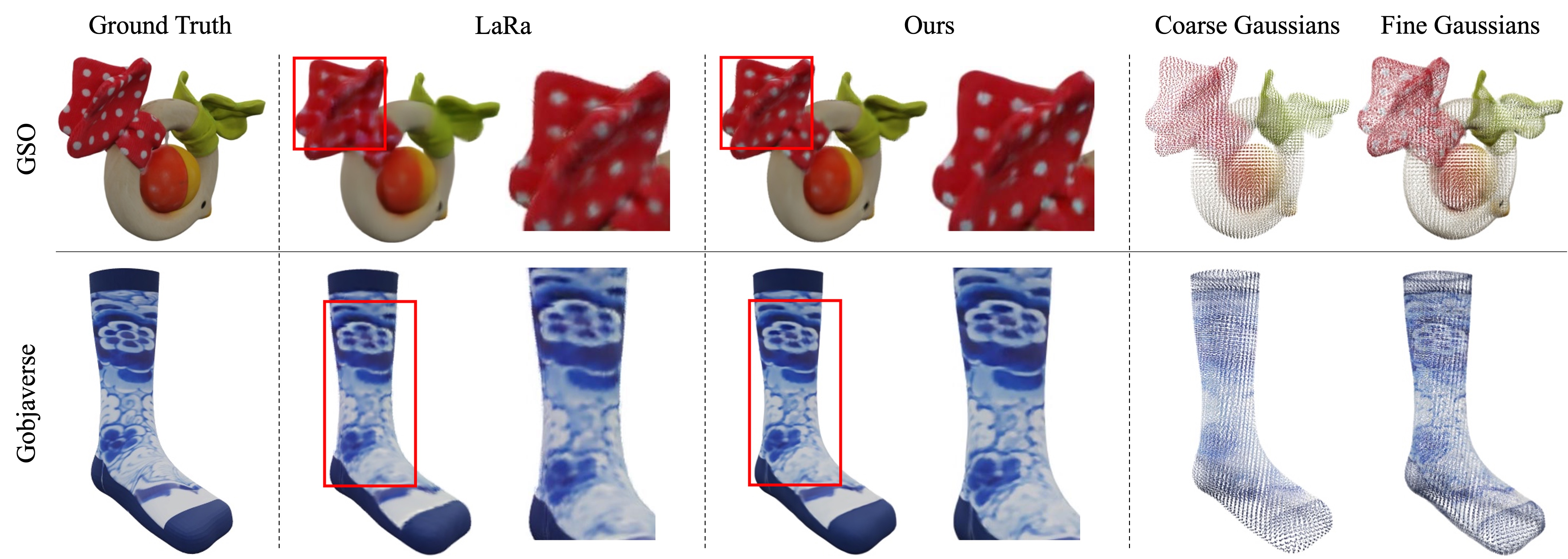}
    \vspace*{-7mm}
    \caption{
    Qualitative comparisons of our object-level model trained for 50 epochs against the original LaRa.
    The zoomed-in parts within the red boxes are shown on the right side of the second and third columns, focusing on the comparison of fine detail reconstruction.
    The two images in the rightmost column present the Gaussians input to and output from our generative densification module, respectively.
    }
    \vspace{-3mm}
    \label{fig:experiment_object}
\end{figure*}

\section{Experiment}
\label{sec:experiment}

\subsection{Experimental Setup}
\label{subsec:settings}

\paragraph{Implementation Details.}
We jointly trained the generative densification module with the LaRa backbone for 30 epochs, and then fine-tuned for 20 epochs to achieve further improvements in rendering quality.
The training was conducted on four A6000-48G GPUs over 3.5 days, with a batch size of 4 per GPU.
Similarly, the MVSplat backbone was trained jointly with our densification module for 300,000 iterations, followed by fine-tuning for 150,000 iterations.
The scene-level training was conducted on four H100-80G GPUs over 6.5 days, also with a batch size of 4 per GPU.
For a detailed description of the network and training hyperparameters, please refer to \cref{sec:suppl_training_and_evaluation_details}.

\vspace{-3mm}
\paragraph{Datasets.}
We train and evaluate our object-level model on Gobjaverse~\cite{xu-gobjaverse} dataset, a large-scale multi-view dataset with an image resolution of 512$\times$512.
To further demonstrate the cross-domain capability, we evaluate our method on Google Scanned Objects~\cite{downs2022google} dataset and a subset of Co3D~\cite{reizenstein2021common} test set, following LaRa~\cite{chen2025lara}. 
Similarly, we train and evaluate our scene-level model on RE10K~\cite{zhou2018stereo} with an image resolution of 256$\times$256, and further evaluate it on two cross-domain datasets, ACID~\cite{infinite_nature_2020} and DTU~\cite{jensen2014large}, following MVSplat~\cite{chen2025mvsplat}.
For both models, we use training and testing splits provided in each dataset without any modifications.

\vspace{-3mm}
\paragraph{Baselines.}
We compare our object-level model against the original LaRa~\cite{chen2025lara}, as well as LGM~\cite{tang2025lgm} and GS-LRM~\cite{zhang2025gs-lrm}.
Additionally, we include two feed-forward NeRF models, MVSNeRF~\cite{chen2021mvsnerf} and MuRF~\cite{xu2024murf}, for reference.
For comparing our scene-level model, we include feed-forward NeRF models (pixelNeRF~\cite{zhang2024pixel}, GNPR~\cite{suhail2022generalizable}, and MuRF~\cite{xu2024murf}) and pixel-aligned Gaussian models (pixelSplat~\cite{charatan2024pixelsplat}, DepthSplat~\cite{xu2024depthsplat}, and the original MVSplat~\cite{chen2025mvsplat}) as baselines.

\subsection{Results}
\label{subsec:results}

\begin{figure*}[t]
    \centering
    \includegraphics[width=1.0\textwidth]{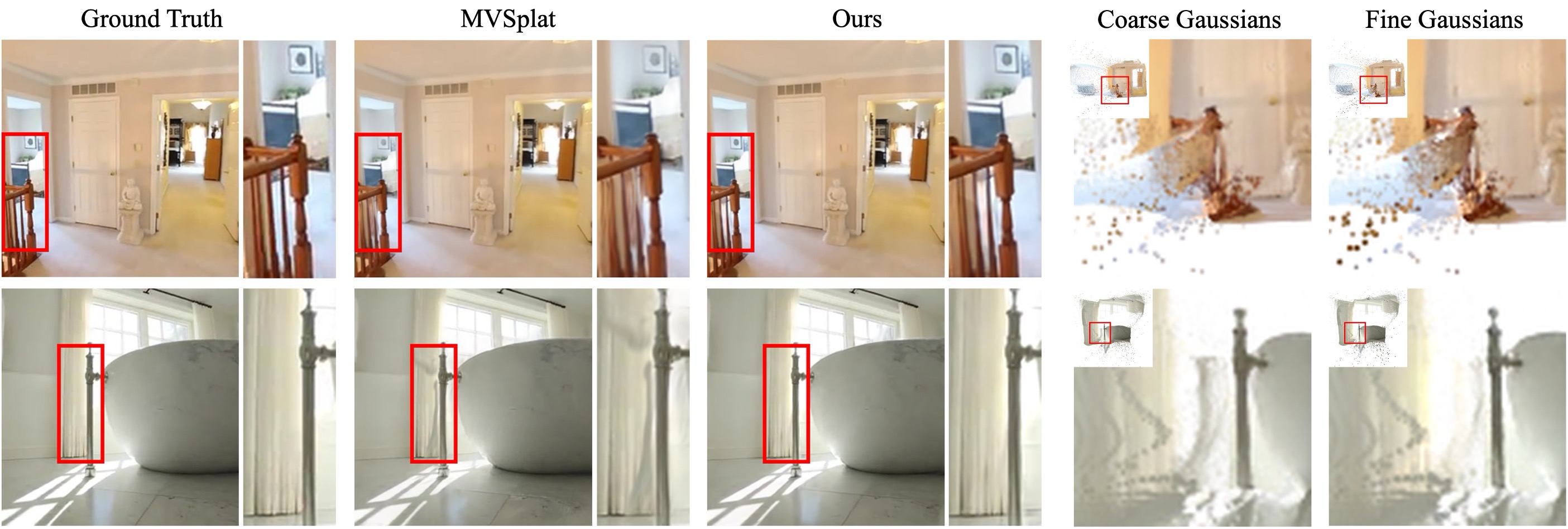}
    \vspace*{-7mm}
    \caption{Qualitative comparisons of our scene-level model against the original MVSplat on the RE10K~\cite{zhou2018stereo} dataset.}
    \vspace{-1mm}
    \label{fig:experiment_scene}
\end{figure*}

\begin{figure*}[t]
    \centering
    \includegraphics[width=1.0\textwidth]{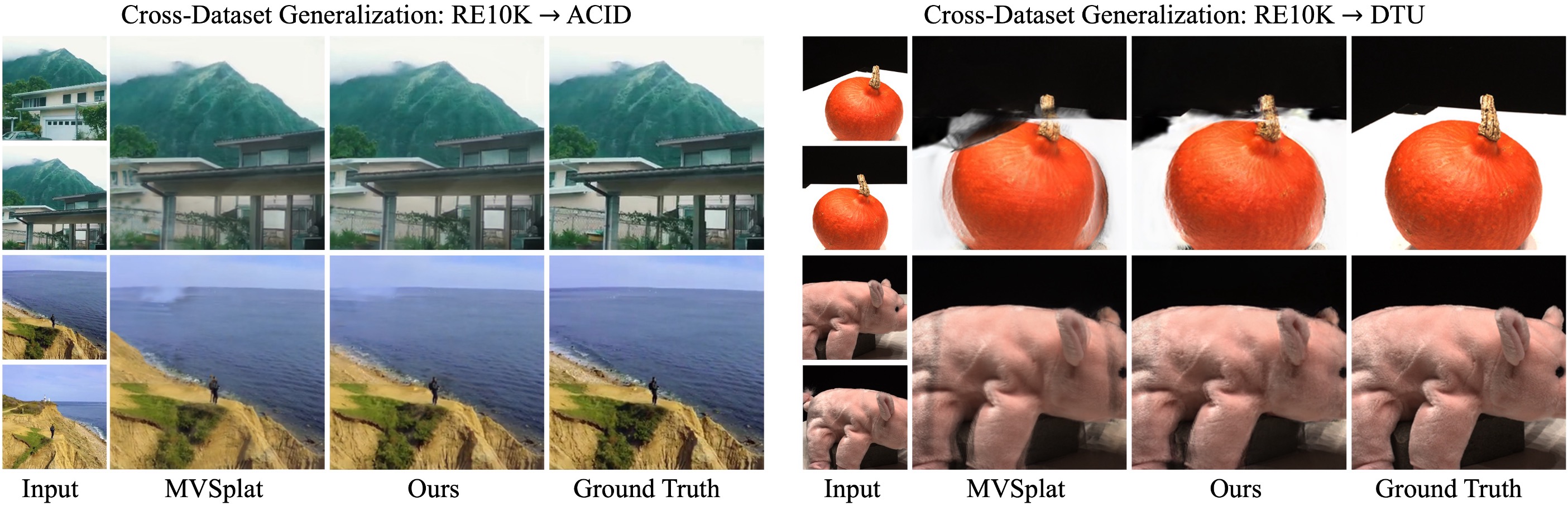}
    \vspace*{-7mm}
    \caption{
    Qualitative comparisons of our scene-level model against the original MVSplat on the ACID~\cite{infinite_nature_2020} and DTU~\cite{jensen2014large} datasets.
    }
    \vspace{-4mm}
    \label{fig:experiment_cross_generalization}
\end{figure*}

\paragraph{Object-level Reconstruction.}
\cref{fig:experiment_object} compares the rendered images from the baseline LaRa against those from our object-level model.
Our model clearly outperforms LaRa in reconstructing fine details, such as the white dots on the red ribbon (first row of \cref{fig:experiment_object}) and the intricate floral patterns on the socks (second row of \cref{fig:experiment_object}).
It is also evident that our densification strategy effectively captures detailed areas and generate fine Gaussians, as shown in the rightmost column of \cref{fig:experiment_object}.
While LaRa quickly learns the object's general shape by leveraging explicit volume representations, it struggles to accurately represent edges or contours. 

The quantitative comparison provided in \cref{tab:quantitative result on object} further demonstrates the ability of our model to capture and reconstruct fine details. 
Our model achieves the highest PSNR in both in-domain reconstruction and cross-dataset generalization tasks. 
It even outperforms GS-LRM, the current state-of-the-art model for object-level reconstruction, using significantly fewer number of parameters (300M vs.\ 134M). 
This highlights that selectively generating Gaussians in the detailed areas can be more effective than uniformly generating numerous Gaussians across the entire object.

\vspace{-3mm}
\paragraph{Scene-level Reconstruction.}
\cref{fig:experiment_scene} and \cref{fig:experiment_cross_generalization} compares the rendered images from the baseline MVSplat against those from our scene-level model.
Our model better reconstructs thin structures and fine details, such as the guardrail (first row of \cref{fig:experiment_scene}) and the wave (second row, first column of \cref{fig:experiment_cross_generalization}).
Moreover, we empirically found that our method reduces the opacity of Gaussians in empty space, making them less visible in the rendered images.
For example, it makes the floaters more transparent near the water pipe (second row of \cref{fig:experiment_scene}) and the iron fence (first row, first column of \cref{fig:experiment_cross_generalization}), removing artifacts in the images.

In the quantitative comparisons, our model not only outperforms the baselines including DepthSplat~\cite{xu2024depthsplat} on the in-domain reconstruction task (\cref{tab:quantitative result on scene}), but also consistently achieves the best performance on the cross-dataset generalization task (\cref{tab:cross-dataset generalization on scene}).
While DepthSplat is concurrent work, we include it as a baseline for the reader's reference.
Our model outperforms DepthSplat with fewer parameters (37M vs.\ 28M) and a smaller batch size (32 vs.\ 16).
Additionally, unlike DepthSplat, which relies on a pretrained depth predictor, our model is solely trained with multi-view images.

\begin{table}[t]
    \centering
    \caption{
    Quantitative comparisons of our scene-level model against its baselines on the RE10K~\cite{zhou2018stereo} dataset.
    The MVSplat-finetune model is fine-tuned for 150,000 iterations from the MVSplat checkpoint at 300,000 iterations.
    * indicates concurrent work.
    }
    \vspace*{-3mm}
    \resizebox{\columnwidth}{!}{
    \small
        \begin{tabular}{lcccc}
        \toprule
        Method & \#Param(M) & PSNR $\uparrow$ & SSIM $\uparrow$ & LPIPS $\downarrow$ \\
        \midrule
        pixelNeRF~\cite{yu2021pixelnerf} & 250 & 20.43 & 0.589 & 0.550 \\
        GPNR~\cite{suhail2022generalizable} & 27 & 24.11 & 0.793 & 0.255 \\
        MuRF~\cite{xu2024murf} & 15.7 & 26.10 & 0.858 & 0.143 \\
        \midrule
        pixelSplat~\cite{charatan2024pixelsplat} & 125.1 & 25.89 & 0.858 & 0.142 \\
        MVSplat~\cite{chen2025mvsplat} & 12.0 & 26.39 & 0.869 & 0.128\\
        MVSplat-finetune & 12.0 & 26.46 & 0.870 & 0.127 \\
        DepthSplat*~\cite{xu2024depthsplat} & 37 & 26.76 & 0.877 & 0.123 \\
        Ours & 27.8 & \textbf{27.08} & \textbf{0.879} & \textbf{0.120} \\
        \bottomrule
        \end{tabular}
    }
    \vspace{-2mm}
    \label{tab:quantitative result on scene}
\end{table}

\begin{table}[t]
    \centering
    \caption{
    Cross-dataset generalization results on the ACID~\cite{infinite_nature_2020} and DTU~\cite{jensen2014large} datasets. 
    All models are trained on the RE10K~\cite{zhou2018stereo} training set and evaluated on each dataset without fine-tuning.
    }
    \vspace*{-3mm}
    \resizebox{\columnwidth}{!}{
    \small
        \begin{tabular}{lcccccc}
        \toprule
        \multirow{2}{*}{Method} & \multicolumn{3}{c}{ACID~\cite{infinite_nature_2020}} & \multicolumn{3}{c}{DTU~\cite{jensen2014large}} \\
        \cmidrule(lr){2-4}\cmidrule(lr){5-7}
         & PSNR $\uparrow$ & SSIM $\uparrow$ & LPIPS $\downarrow$ & PSNR $\uparrow$ & SSIM $\uparrow$ & LPIPS $\downarrow$ \\
        \midrule
        pixelSplat~\cite{charatan2024pixelsplat} & 27.64 & 0.830 & 0.160 & 12.89 & 0.382 & 0.560 \\
        MVSplat~\cite{chen2025mvsplat} & 28.15 & 0.841 & 0.147 & 13.94 & 0.473 & 0.385 \\
        Ours & \textbf{28.61} & \textbf{0.847} & \textbf{0.141} & \textbf{14.05} & \textbf{0.477} & \textbf{0.380} \\
        \bottomrule
        \end{tabular}
    }
    \vspace{-2mm}
    \label{tab:cross-dataset generalization on scene}
\end{table}

\begin{table}
    \centering
    \caption{
    Ablations on the number of selected Gaussians and learnable masking, evaluated on the GSO~\cite{downs2022google} dataset.
    Mask and $K^{(0)}$ denote learnable masking and the number of selected Gaussians in gradient masking, respectively.
    \#Gausss refers to the number of final Gaussians, the union of the coarse and fine Gaussians.
    }
    \vspace*{-3mm}
    \resizebox{\columnwidth}{!}{
    \small
        \begin{tabular}{cc|cccc}
        \toprule
        Mask & $K^{(0)}$ & \#Gauss $\downarrow$ & PSNR $\uparrow$ & SSIM $\uparrow$ & LPIPS $\downarrow$ \\
        \midrule
        \checkmark & 12,000 & 114,351 & 30.62 & 0.964 & 0.062 \\
        \midrule
        \checkmark & 0 & \textbf{46,693} & 29.46 & 0.951 & 0.079 \\
        \checkmark & 15,000 & 130,056 & 30.68 & 0.964 & 0.061 \\
        \checkmark & 30,000 & 189,322 & 30.63 & 0.964 & 0.061 \\
        & 12,000 & 151,968 & \textbf{30.73} & \textbf{0.965} & \textbf{0.060} \\
        \bottomrule
        \end{tabular}
    }
    \vspace{-5mm}
    \label{tab:ablation study}
\end{table}

\subsection{Comparing Coarse and Fine Gaussians}
\label{subsec:comparing_coarse_and_fine_gaussians}
We provide an intuitive example comparing coarse and fine Gaussians, with an analysis of their positions, opacities, and scales.
Here, we refer coarse Gaussians to the input Gaussians selected by the gradient masking, while fine Gaussians to their corresponding generated fine Gaussians (i.e., the union of output Gaussians from each densification layer).

\begin{figure}[t]
    \centering
    \includegraphics[width=1.0\columnwidth]{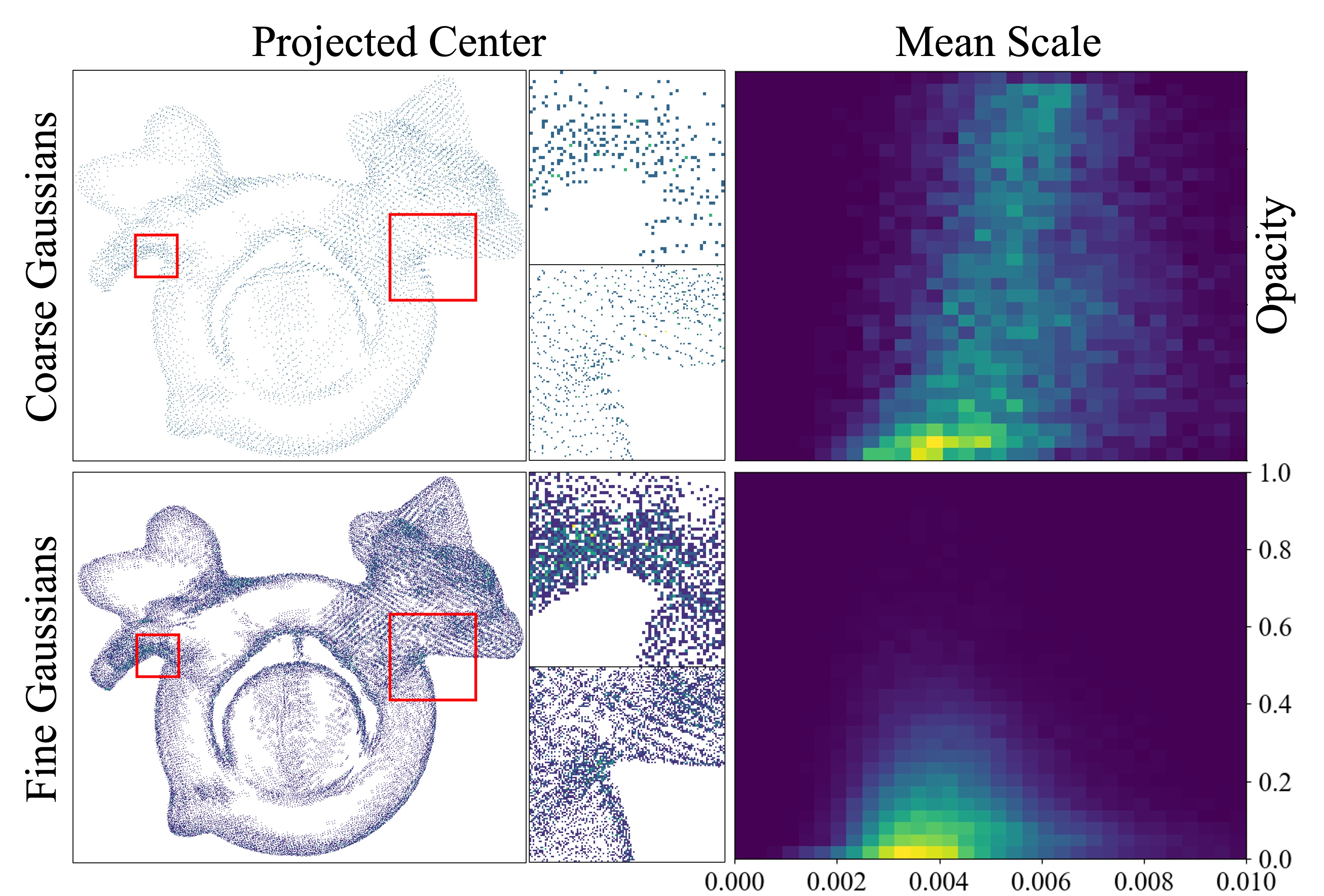}
    \vspace*{-7mm}
    \caption{
    2D histograms of Gaussian attributes.
    Each pixel represents a histogram bin, with brighter colors for higher counts.
    }
    \vspace{-2mm}
    \label{fig:analysis_histogram}
\end{figure}

The left side of \cref{fig:analysis_histogram} illustrates that reconstructing contours is challenging with insufficient number of Gaussians.
Although these areas are covered by large Gaussians with no visible holes in the rendered image, it still appears blurry, with the object not clearly separated from the background (\cref{fig:experiment_object}).
As shown on the right side, fine Gaussians have smaller scales and lower opacities compared to coarse Gaussians, indicating that our method reconstructs details by accumulating partially overlapping small Gaussians.

\subsection{Ablation Study}
\label{subsec:ablation_study}

\cref{tab:ablation study} presents the ablation results on the number of input Gaussians and learnable masking.
The first row shows the evaluation of our object-level model on the GSO~\cite{downs2022google} dataset, which is identical to the `Ours-fast' model in \cref{tab:quantitative result on object}.
The second to fourth rows show the evaluation of the same model without fine-tuning, varying the number of Gaussians selected by the gradient masking module for densification.
The last row shows the evaluation of our object-level model retrained for 30 epochs without learnable masking.

\paragraph{Gradient Masking.}
Image quality improves as the number of selected Gaussians is increased from 0 to 12,000 to 15,000 but does not improve further when increased to 30,000 (\cref{tab:ablation study}), indicating that simply densifying more Gaussians does not guarantee better image quality.
Moreover, the number of final Gaussians (\#Gauss) increases by approximately 66\% when $K^{(0)}$ is increased from 12,000 (our default setting) to 30,000, which leads to slower rendering and higher GPU memory requirements in practice.
Our densification strategy balances image quality and rendering efficiency by using the gradient masking to selectively densifying only the necessary Gaussians.

\vspace{-3mm}
\paragraph{Learnable Masking.}
Learnable masking is another crucial component for reducing the computational and memory demands, as it controls the number of Gaussians generated by each densification layer.
Ideally, the masking should filter out Gaussians that do not require further densification without compromising image quality.
As shown in the last row of \cref{tab:ablation study}, although the masking results in a slight decrease in PSNR of 0.36\%, it significantly reduces the number of final Gaussians by 25\%.
This demonstrates that the computational and memory efficiency gained by reducing the number of Gaussians outweighs the minor loss in image quality, highlighting the importance of learnable masking.

\section{Conclusion}
\label{sec:conclution}

We proposed generative densification, an efficient densification strategy for generalized feed-forward models.
We integrated the proposed method into LaRa and MVSplat, providing practical guidance for real-world applications.
Extensive experiments demonstrate that our method generates fine Gaussians capable of reconstructing high-frequency details, establishing new benchmarks in both object-level and scene-level reconstruction tasks.
We believe our work opens a new research direction towards generating fine Gaussians for high-fidelity generalizable 3D reconstruction. 

{
    \small
    \bibliographystyle{ieeenat_fullname}
    \bibliography{main}
}

\clearpage
\appendix
\renewcommand{\thesection}{\Alph{section}} 
\section*{Appendix}

\section{Additional Results on the DL3DV Dataset}
\label{sec:suppl_additional_results_dl3dv}

\begin{table}[h]
    \centering
    \caption{
    Evaluation results on the DL3DV-10K~\cite{ling2024dl3dv} dataset.
    $\boldsymbol{n}$ denotes the frame distance span across all test views
    }
    \vspace*{-3mm}
    \resizebox{\columnwidth}{!}{
        \begin{tabular}{ccccccc}
        \toprule
        \multirow{2}{*}{\textbf{Method}} & \multicolumn{3}{c}{$\boldsymbol{n=150}$} & \multicolumn{3}{c}{$\boldsymbol{n=300}$} \\
        \cmidrule(lr){2-4}\cmidrule(lr){5-7}
        & PSNR $\uparrow$ & SSIM $\uparrow$ & LPIPS $\downarrow$ & PSNR $\uparrow$ & SSIM $\uparrow$ & LPIPS $\downarrow$ \\
        \midrule
        MVSplat-finetune & 17.42 & 0.516 & 0.417 & 16.19 & 0.452 & 0.478 \\
        Ours & \textbf{17.76} & \textbf{0.533} & \textbf{0.405} & \textbf{16.46} & \textbf{0.468} & \textbf{0.469} \\
        \bottomrule
        \end{tabular}
    }
    \vspace{-0.5em}
    \label{tab:suppl_additional_results_dl3dv}
\end{table}

\noindent We conducted an additional evaluation on the DL3DV-10K~\cite{ling2024dl3dv}, a challenging dataset with 51.3 million frames from 10,510 real-world scenes.
Our scene-level model and its baseline are fine-tuned for 100,000 iterations on two subsets (``3K" and ``4K") of the DL3DV-10K dataset, comprising approximately 2,000 scenes.
All models are evaluated on 140 scenes that are filtered out from the training set, following MVSplat360~\cite{chen2024mvsplat360}.
For each scene, we select 5 views as input and evaluate on 56 views uniformly sampled from the remaining views~\cite{chen2024mvsplat360}.
As shown in \cref{tab:suppl_additional_results_dl3dv}, our model outperforms the baseline across all metrics, consistent with the evaluations on the RE10K, ACID, and DTU datasets.

\section{Generating Residuals of Fine Gaussians}
\label{sec:suppl_generating_residuals}

We further propose to generate residuals of fine Gaussians for scene-level reconstruction, which involves capturing complex geometries and appearances across diverse indoor and outdoor environments.

Similar to the densification procedure presented in \cref{subsec:generative densification}, the top $K^{(0)}$ input Gaussians with large view-space positional gradients ($\mathcal{G}^{(0)}_\text{den}$) are selected, and their positions and features ($\mathcal{X}_\text{den}^{(0)}$, $\mathcal{F}_\text{den}^{(0)}$) are passed through alternating layers of up-sampling ($\texttt{UP}$), Gaussian head ($\texttt{HEAD}$), and splitting ($\texttt{SPLIT}$) modules:
\begin{align}
    (\mathcal{X}^{(l)}, \mathcal{F}^{(l)}) &= \texttt{UP}(\mathcal{X}_\text{den}^{(l-1)}, \mathcal{F}_\text{den}^{(l-1)}), \label{eq:eq12} \\
    \mathcal{G}^{(l)} &= \texttt{HEAD}(\mathcal{G}_\text{den}^{(l-1)}, \mathcal{X}^{(l)}, \mathcal{F}^{(l)}), \label{eq:eq13} \\
    (\mathcal{G}_\text{den}^{(l)}, \mathcal{X}_\text{den}^{(l)}, \mathcal{F}_\text{den}^{(l)}, \mathcal{G}_\text{rem}^{(l)}) &= \texttt{SPLIT}(\mathcal{G}^{(l)}, \mathcal{X}^{(l)}, \mathcal{F}^{(l)}), \label{eq:eq14}
\end{align}
for $l \in \{1, {\cdots}, L{-}1\}$.
The Gaussian positions and features are up-sampled in the $\texttt{UP}$ module, which is identical to the up-sampling procedure as described in \cref{subsec:architecture_details}.
The up-sampled positions and features are then passed to the $\texttt{HEAD}$ module, where they are transformed into fine Gaussian parameters and added to those from the previous layer ($\mathcal{G}_\text{den}^{(l-1)}$).
In the $\texttt{SPLIT}$ module, the fine Gaussians are divided into two groups: those that require further densification and those do not.
The first group of Gaussians is refined in the next layer, while the second group remains unchanged.
Note that the second group of Gaussians' positions and features ($\mathcal{X}_\text{rem}^{(l)}, \mathcal{F}_\text{rem}^{(l)}$) are omitted from \cref{eq:eq14} for brevity.
Finally, the $L$-th layer's fine Gaussians are generated as $\mathcal{G}^{(L)} = \texttt{HEAD}(\mathcal{G}_\text{den}^{(L-1)}, \texttt{UP}(\mathcal{X}_\text{den}^{(L-1)}, \mathcal{F}_\text{den}^{(L-1)}))$ without splitting, and the final set of Gaussians is obtained as:
\begin{equation}
\label{eq:eq15}
    \hat{\mathcal{G}} = \{ \bigcup_{l=0}^{L-1} \mathcal{G}_\text{rem}^{(l)} \} \cup \mathcal{G}^{(L)}.
\end{equation}

The two densification methods for object-level and scene-level reconstruction are similar in that both selectively generates fine Gaussians leveraging learnable masking, but they differ in how fine Gaussians are generated.
The object-level method generates fine Gaussians directly in each densification layer, while the scene-level method generates initial fine Gaussians in the first layer and selectively refines them by adding residuals in subsequent layers.

Although the residual learning is proposed for the scene-level model, it can also be applied to the object-level model.
As shown in \cref{tab:quantitative result on object}, our object-level model trained with the residual learning achieves the best PSNR, demonstrating its effectiveness in object-level reconstruction.

\section{Model Details}
\label{sec:suppl_model_details}

\paragraph{The Number of Gaussians.}
We set $K^{(0)}$, the number of selected Gaussians in the gradient masking module, to 12,000 for object-level reconstruction and 30,000 for scene-level reconstruction.
The selected Gaussian positions and features are up-sampled by a factor of $R^{(l)}$, leading to an increased number of Gaussians given by $K^{(l)}=K^{(l-1)}R^{(l)}$.
In the splitting module, the up-sampled Gaussians are divided into two groups: those that need another round of densification in the next layer and those do not.
The number of Gaussians for further densification and the remaining ones are defined as $K^{(l)}_\text{den}=\lceil K^{(l)}P^{(l)} \rceil$ and $K^{(l)}_\text{rem}=K^{(l)}-K^{(l)}_\text{den}$, respectively, where $P^{(l)}\in(0,1)$ denotes the masking ratio and $\lceil\cdot\rceil$ is the ceiling operator.

\paragraph{Generative Densification of LaRa.}
LaRa~\cite{chen2025lara} generates 3D volume representations conditioned on image features, and each volume features are decoded into multiple Gaussians.
To improve the rendering quality, LaRa introduce a cross-attention between the volume features and coarse renderings, including ground-truth images, rendered images, depth images, and accumulated alpha maps~\cite{chen2025lara}.
The intermediate features from the cross-attention are transformed to residuals of SH using an MLP, which are added to the coarse SH to obtain the refined Gaussians.

We modify the last MLP (the residual SH decoder) to output both the residuals and refined volume features.
The first $d$ columns of the MLP output are considered the residual SH, while the remaining columns serve as input features for generative densification. 
Here, $d$ denotes the number of SH coefficients.
We take the refined Gaussians and the concatenation of volume and refined volume features as input, and their respective fine Gaussians are generated by our method.
Note that, while the baseline LaRa generates 2D Gaussian representations~\cite{huang20242d}, we adapt it to generate 3D Gaussians representations~\cite{kerbl20233d} instead.

\paragraph{Generative Densification of MVSplat.}
MVSplat~\cite{chen2025mvsplat} generates per-view pixel-aligned Gaussian representations from input multi-view images.
A transformer encodes the input images into features via cross-view attention, after which per-view cost volumes are constructed.
The image features are concatenated with these cost volumes and decoded into depths and other parameters, including opacities, covariances, and colors.
The Gaussian positions are then determined by un-projecting the depths into 3D space.

Similar to LaRa, we obtain the refined features by applying cross-attention between the coarse renderings and the concatenated features of images and cost volumes, followed by a simple MLP.
However, we do not predict residuals of SH, as this often leads to unstable training in scene-level reconstruction.
We use the per-view Gaussians from the MVSplat backbone along with the refined features as input to our method, generating per-view fine Gaussians.

\paragraph{Impelmentation Details.}
For object-level reconstruction, we select 12,000 Gaussians from the LaRa backbone with large view-space gradients and generate fine Gaussians through two densification layers.
The up-sampling factors of the two layers are 2 and 4, and the masking ratio is 0.8.
In other words, the input Gaussians are densified by a factor of 2 in the first layer, and 80\% of them are further densified by a factor of 4 in the second layer, while the remaining 20\% are decoded into raw Gaussians.
Similarly, for scene-level reconstruction, we select 30,000 Gaussians per view from the MVSplat backbone.
We use three densification layers, each with an up-sampling factor of 2.
The masking ratios are set to 0.5 for the first layer and 0.8 for the second layer.

\section{Training and Evaluation Details}
\label{sec:suppl_training_and_evaluation_details}
The training and fine-tuning hyperparameters are summarized in \cref{tab:training_config} and \cref{tab:finetuning_config}, respectively.
The training objectives and additional details are outlined in the followings.

\paragraph{Object-level Reconstruction.}
The backbone and the densification module are jointly trained by minimizing the loss $\mathcal{L} = \mathcal{L}_\text{MSE}(\mathcal{I}, \hat{\mathcal{I}}) + 0.5 (1 - \mathcal{L}_\text{SSIM}(\mathcal{I}, \hat{\mathcal{I}}))$, for both coarse and fine images, where $\mathcal{L}_\text{MSE}$ is the mean squared error and $\mathcal{L}_\text{SSIM}(\mathcal{I}, \hat{\mathcal{I}})$ is the structural similarity loss.
The coarse images are rendered using the Gaussians generated by the backbone LaRa, and the fine images are rendered using the final Gaussians (\cref{eq:eq7}) from our densification module.
Unlike the original implementation~\cite{chen2025lara}, where the fine decoder (the last cross-attention layer and the residual SH decoder) is trained after the first 5,000 iterations, we train it from the very beginning.

For model evaluation on the GSO dataset, we utilize the classical K-means algorithm to group the cameras into 4 clusters and select the center of each cluster to ensure sufficient angular coverage of the input views.
Both LaRa and our model are evaluated using this new sampling method.

\paragraph{Scene-level Reconstruction.}
Similar to our object-level model, we calculate the image reconstruction loss $\mathcal{L}=L_\text{MSE}(\mathcal{I}, \hat{\mathcal{I}}) + 0.05\mathcal{L}_\text{LPIPS}(\mathcal{I}, \hat{\mathcal{I}})$ for both coarse and fine images, and minimize the loss to jointly train the MVSplat backbone and the densification module.
Here, $\mathcal{L}_\text{LPIPS}$ denotes the learned perceptual similarity loss (LPIPS~\cite{zhang2018unreasonable}).
For model fine-tuning, we set the warm-up step of the view-sampler to 0, and the minimum and maximum distances between context views are set to 45 and 192, respectively.

\begin{table}
    \small
    \centering
    \caption{Summary of training hyperparameters.}
    \vspace*{-3mm}
    \resizebox{\columnwidth}{!}{
        \begin{tabular}{lcclc}
        \toprule
        \multicolumn{2}{c}{Object-level} & & \multicolumn{2}{c}{Scene-level} \\
        \cline{1-2} \cline{4-5}
        Config & \multicolumn{1}{c}{Value} & & Config & \multicolumn{1}{c}{Value} \\
        \midrule
        optimizer & AdamW~\cite{loshchilov2017decoupled} & & optimizer & Adam~\cite{kingma2014adam} \\
        scheduler & Cosine & & scheduler & Cosine \\
        learning rate & 4e-4 & & learning rate & 2e-4 \\
        beta & [0.9, 0.95] & & beta & [0.9, 0.999] \\
        weight decay & 0.05 & & weight decay & 0.00 \\
        warmup iters & 1,000 & & warmup iters & 2,000 \\
        epochs & 30 & & iters & 300,000 \\
        \bottomrule
        \end{tabular}
    }
    \label{tab:training_config}
\end{table}

\begin{table}
    \small
    \centering
    \caption{Summary of fine-tuning hyperparameters.}
    \vspace*{-3mm}
    \resizebox{\columnwidth}{!}{
        \begin{tabular}{lcclc}
        \toprule
        \multicolumn{2}{c}{Object-level} & & \multicolumn{2}{c}{Scene-level} \\
        \cline{1-2} \cline{4-5}
        Config & \multicolumn{1}{c}{Value} & & Config & \multicolumn{1}{c}{Value} \\
        \midrule
        optimizer & AdamW~\cite{loshchilov2017decoupled} & & optimizer & Adam~\cite{kingma2014adam} \\
        scheduler & Cosine & & scheduler & Cosine \\
        learning rate & 2e-4 & & learning rate & 2e-4 \\
        beta & [0.9, 0.95] & & beta & [0.9, 0.999] \\
        weight decay & 0.05 & & weight decay & 0.00 \\
        warmup iters & 0 & & warmup iters & 0 \\
        epochs & 20 & & iters & 150,000 \\
        \bottomrule
        \end{tabular}
    }
    \label{tab:finetuning_config}
\end{table}

\section{Additional Qualitative Results}
\label{sec:additional_qualitative_comparisons}

\begin{figure*}[t]
    \centering
    \includegraphics[width=1.0\textwidth]{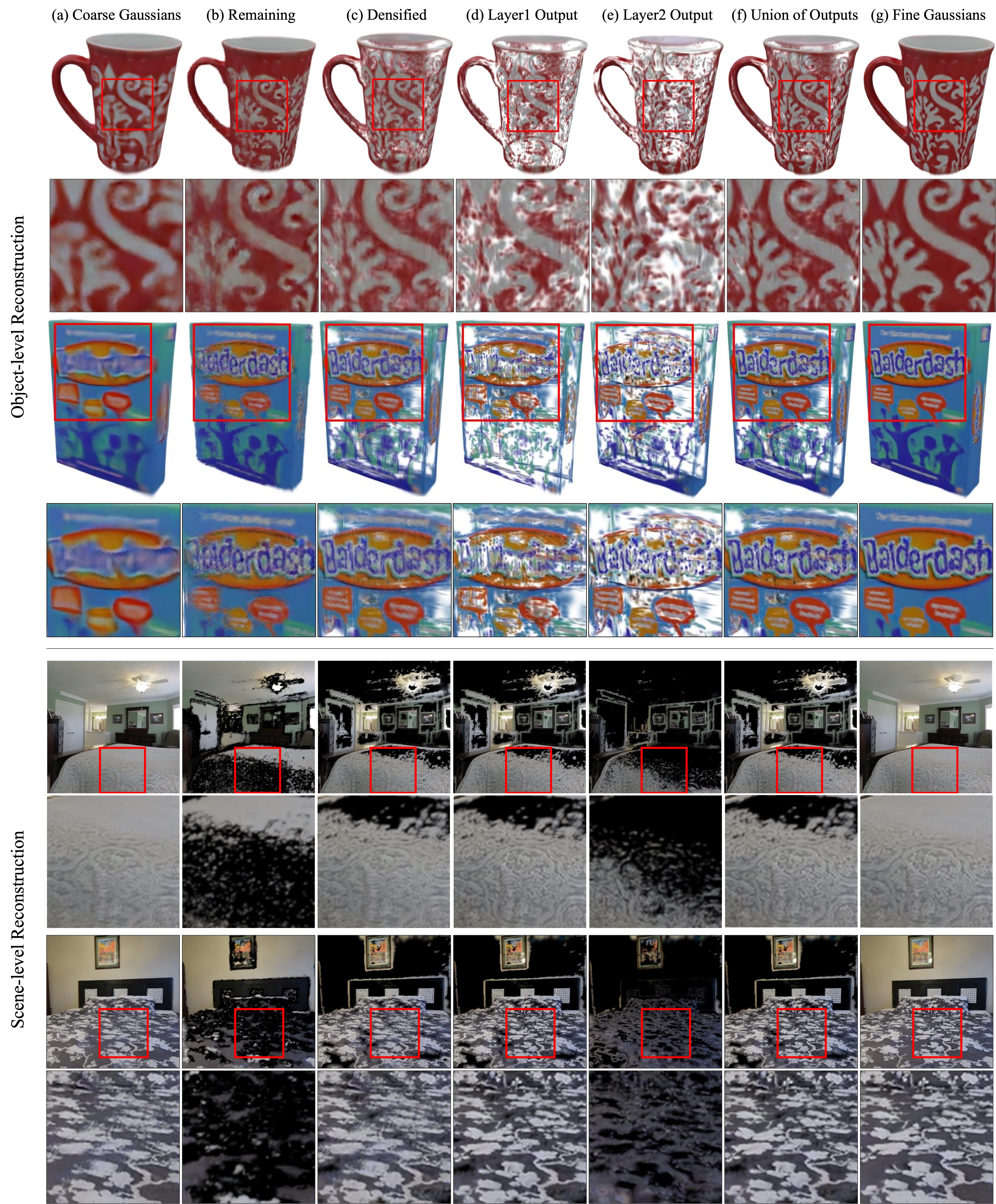}
    \caption{
    Additional qualitative results of our object-level and scene-level model trained for 50 epochs and 450,000 iterations, respectively. The zoomed-in parts show our method selects and reconstructs the fine details through alternating densification layers, while preserving the smooth areas unchanged. Note that the 7-th column of the scene-level reconstruction results shows the union of fine Gaussians generated across all three densification layers, and the output Gaussians from the third layer are omitted due to space constraints.
    }
    \label{fig:suppl_main}
\end{figure*}

\begin{figure*}[t]
    \centering
    \includegraphics[width=1.0\textwidth]{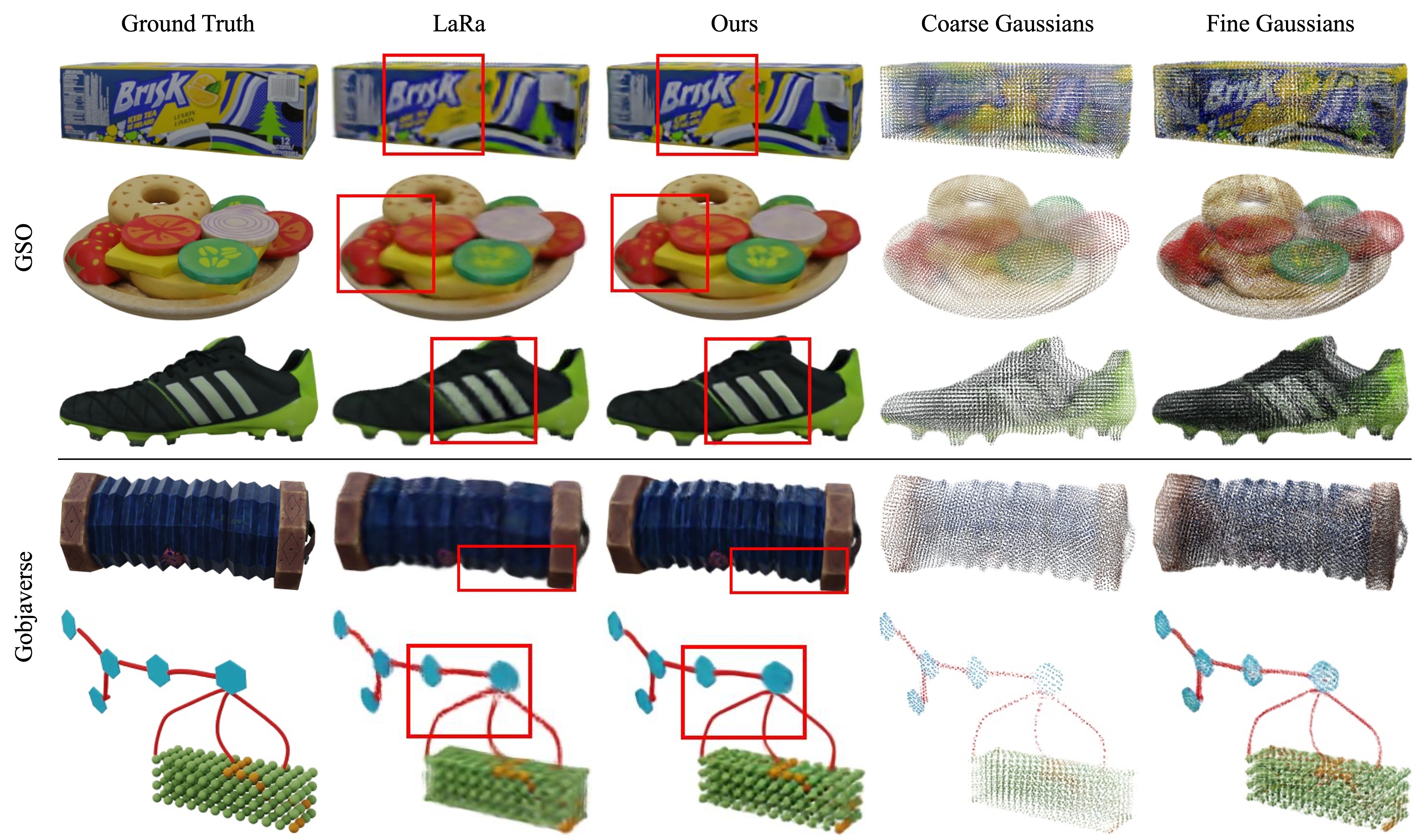}
    \vspace*{-6mm}
    \caption{
    Qualitative comparisons of our object-level model against the original LaRa~\cite{chen2025lara}, evaluated on the GSO~\cite{downs2022google} and Gobjaverse~\cite{xu-gobjaverse} dataset. The coarse and fine Gaussians are the input and output of generative densification module, respectively.
    }
    \vspace{1.0em}
    \label{fig:suppl_object}
\end{figure*}

\begin{figure*}[t]
    \centering
    \includegraphics[width=1.0\textwidth]{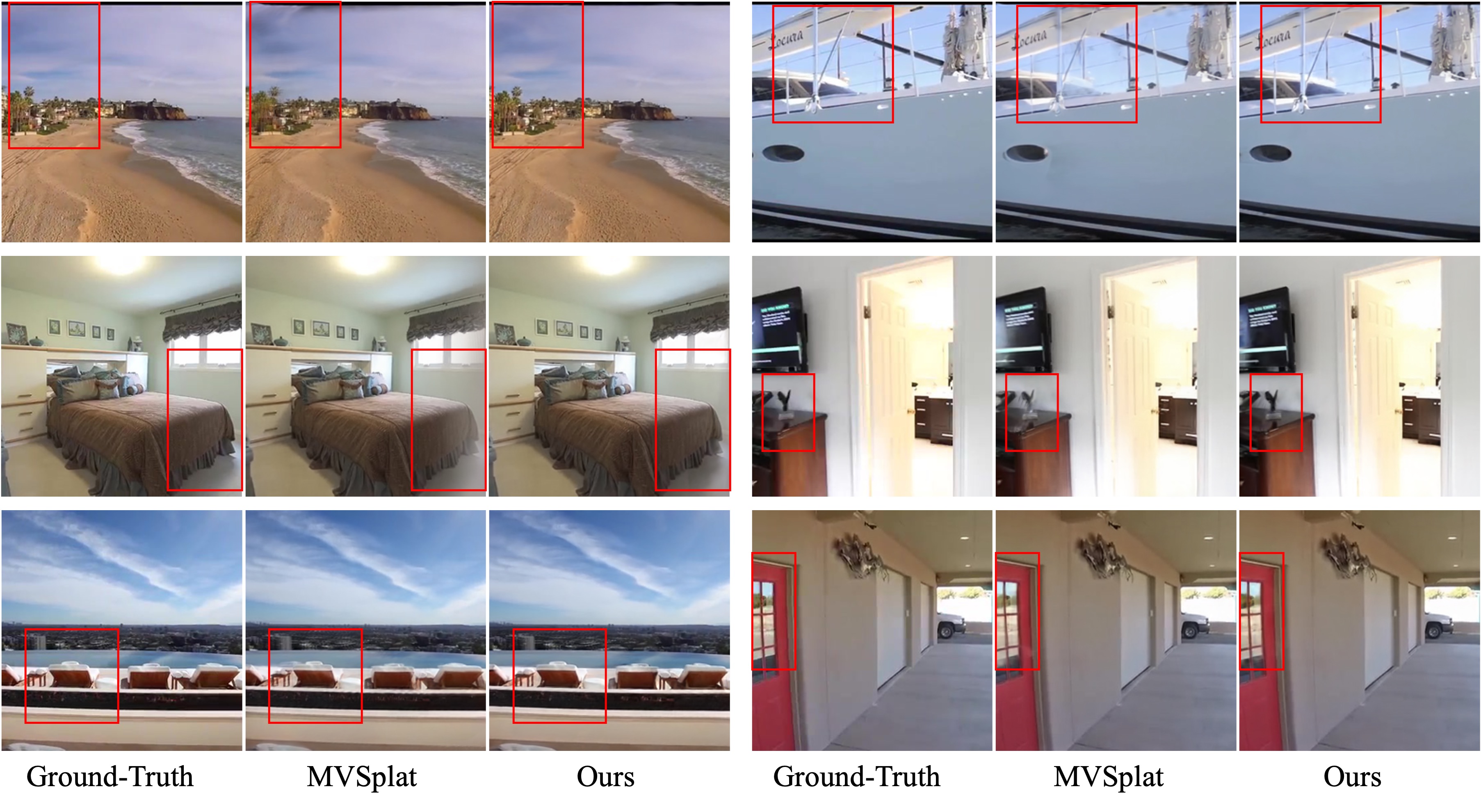}
    \vspace*{-7mm}
    \caption{
    Qualitative comparisons of our scene-level model against the original MVSplat~\cite{chen2025mvsplat}, evaluated on the RE10K~\cite{zhou2018stereo} dataset. The red boxes show that our model better reconstructs the scene, removing visual artifacts and generating missing parts.
    }
    \label{fig:suppl_scene}
\end{figure*}

We provide additional qualitative results for both object-level and scene-level reconstruction tasks.
\cref{fig:suppl_main} illustrates how fine Gaussians are generated in each densification layer.
\cref{fig:suppl_object} and \cref{fig:suppl_scene} show qualitative comparisons for object-level and scene-level reconstruction, respectively.

\end{document}